\documentstyle[eepic,epsf,eaclap]{article}

  \makeatletter

\@ifundefined{new@fontshape}{\def\reset@font{}\let\mathrm\rm}{}

\let\prmbrs=0
\def\primebars{\let\prmbrs=1}
\def\obar#1{\ifmmode#1^{0}\else#1$^{0}$\fi}  
\def\mbar#1{\ifmmode#1^{\mathrm{max}}\else#1$^{\mathrm{max}}$\fi}
\def\ibar#1{\ifx\prmbrs0%
                 \ifmmode\overline{\mathrm{#1}}\else$\overline{\mbox{#1}}$\fi%
            \else\ifmmode#1^{'}\else#1$^{'}$\fi\fi}
\def\iibar#1{\ifx\prmbrs0%
                  \ifmmode\overline{\overline{\mathrm{#1}}}%
                  \else$\overline{\overline{\mbox{#1}}}$\fi%
             \else #1P\fi}
\def\th{\ifmmode\theta\else$\theta$\fi}
\def\al{\ifmmode\alpha\else$\alpha$\fi}
\def\be{\ifmmode\beta\else$\beta$\fi}
\def\ga{\ifmmode\gamma\else$\gamma$\fi}
\def\de{\ifmmode\delta\else$\delta$\fi}

\catcode`_=\active
\def_#1{\ifmmode\mit{\sb{#1}}\else${}\sb{#1}$\fi}
\catcode`^=\active
\def^#1{\ifmmode\mit{\sp{#1}}\else${}\sp{#1}$\fi}

\def\lb#1{\@ifnextchar [{\@glarph{#1}}{\@bl{#1}}}
\def\@glarph#1[#2]{\ifmmode{[}\sb{{\mathrm{#1}}\sb{#2}}\else%
                           ${[}\sb{{\mathrm{#1}}\sb{#2}}$\fi}
\def\@bl#1{\ifmmode{[}\sb{\mathrm{#1}}\;\else${[}\sb{\mathrm{#1}}\;$\fi}

\def\rb#1{\@ifnextchar [{\@grarph{#1}}{\@br{#1}}}
\def\@grarph#1[#2]{\ifmmode{]}\sb{{\mathrm{#1}}\sb{#2}}\else%
                        ${]}\sb{{\mathrm{#1}}\sb{#2}}$\fi}
\def\@br#1{\ifmmode{]}\sb{\mathrm{#1}}\;\else${]}\sb{\mathrm{#1}}\;$\fi}

\def\qlist{\begin{list}{\Alph{xnum}.}{\usecounter{xnum}%
\setlength{\rightmargin}{\leftmargin}}}
\def\endqlist{\end{list}}

\newif\if@noftnote\@noftnotetrue
\newif\if@xrec\@xrecfalse
\@definecounter{fnx}

\long\def\@footnotetext#1{%
    \@noftnotefalse\setcounter{fnx}{0}%
    \insert\footins{\reset@font\footnotesize
    \interlinepenalty\interfootnotelinepenalty
    \splittopskip\footnotesep
    \splitmaxdepth \dp\strutbox \floatingpenalty \@MM
    \hsize\columnwidth \@parboxrestore
   \edef\@currentlabel{\csname p@footnote\endcsname\@thefnmark}\@makefntext
    {\rule{\z@}{\footnotesep}\ignorespaces
      #1\strut}}\@noftnotetrue}

\newcount\@xnumdepth \@xnumdepth = 0

\@definecounter{xnumi}
\@definecounter{xnumii}
\@definecounter{xnumiii}
\@definecounter{xnumiv}
\@definecounter{exx}
\def\thexnumi{\@xsi{xnumi}}
\def\thexnumii{\@xsii{xnumii}}
\def\thexnumiii{\@xsiii{xnumiii}}
\def\thexnumiv{\@xsiv{xnumiv}}
\def\p@xnumii{\thexnumi}
\def\p@xnumiii{\thexnumi\thexnumii-}
\def\p@xnumiv{\thexnumi\thexnumii-\thexnumiii-}

\def\xs@default#1{\csname @@xs#1\endcsname}
\def\@@xsi{\let\@xsi\arabic}
\def\@@xsii{\let\@xsii\alph}
\def\@@xsiii{\let\@xsiii\roman}
\def\@@xsiv{\let\@xsi\arabic}

\@definecounter{rxnumi}
\@definecounter{rxnumii}
\@definecounter{rxnumiii}
\@definecounter{rxnumiv}

\def\save@counters{%
\setcounter{rxnumi}{\value{xnumi}}%
\setcounter{rxnumii}{\value{xnumii}}%
\setcounter{rxnumiii}{\value{xnumiii}}%
\setcounter{rxnumiv}{\value{xnumiv}}}%

\def\reset@counters{%
\setcounter{xnumi}{\value{rxnumi}}%
\setcounter{xnumii}{\value{rxnumii}}%
\setcounter{xnumiii}{\value{rxnumiii}}%
\setcounter{xnumiv}{\value{rxnumiv}}}%

\def\exewidth#1{\def\@exwidth{#1}} \exewidth{(234)}
\def\exe{\@ifnextchar [{\@exe}{\@exe[\@exwidth]}}

\def\@exe[#1]{\ifnum \@xnumdepth >0%
                 \if@xrec\@exrecwarn\fi%
                 \if@noftnote\@exrecwarn\fi%
                 \@xnumdepth0\@listdepth0\@xrectrue%
                 \save@counters%
              \fi%
                 \advance\@xnumdepth \@ne \@@xsi%
                 \begin{list}{(\thexnumi)}%
                             {\usecounter{xnumi}\@subex{#1}{1em}%
                              \if@noftnote%
                                 \setcounter{xnumi}{\value{exx}}%
                              \else%
                                 \setcounter{xnumi}{\value{fnx}}%
                              \fi}}

\def\endexe{\if@noftnote\setcounter{exx}{\value{xnumi}}%
                   \else\setcounter{fnx}{\value{xnumi}}%
                        \reset@counters\@xrecfalse\fi\end{list}}

\def\@exrecwarn{\typeout{*** Recursion on "exe"---your
                example numbering will probably be screwed up!}}

\def\xlist{\@ifnextchar [{\@xlist{}}{\@xlist{}[iv.]}}
\def\xlista{\@ifnextchar [{\@xlist{\alph}}{\@xlist{\alph}[m.]}}
\def\xlisti{\@ifnextchar [{\@xlist{\roman}}{\@xlist{\roman}[iv.]}}
\def\xlistn{\@ifnextchar [{\@xlist{\arabic}}{\@xlist{\arabic}[9.]}}
\def\xlistA{\@ifnextchar [{\@xlist{\Alph}}{\@xlist{\Alph}[M.]}}
\def\xlistI{\@ifnextchar [{\@xlist{\Roman}}{\@xlist{\Roman}[IV.]}}

\def\endxlist{\end{list}}
\def\endxlista{\end{list}}
\def\endxlistn{\end{list}}
\def\endxlistA{\end{list}}
\def\endxlistI{\end{list}}
\def\endxlisti{\end{list}}

\def\@xlist#1[#2]{\ifnum \@xnumdepth >3 \@toodeep\else%
    \advance\@xnumdepth \@ne%
    \edef\@xnumctr{xnum\romannumeral\the\@xnumdepth}%
    \def\@bla{#1}
    \ifx\@bla\empty\xs@default{\romannumeral\the\@xnumdepth}\else%
      \expandafter\let\csname @xs\romannumeral\the\@xnumdepth\endcsname#1\fi
    \begin{list}{\csname the\@xnumctr\endcsname.}%
                {\usecounter{\@xnumctr}\@subex{#2}{1.5ex}}\fi}

\def\@subex#1#2{\settowidth{\labelwidth}{#1}\itemindent\z@\labelsep#2%
         \ifnum\the\@xnumdepth=1\topsep 7\p@ plus2\p@ minus3\p@\else%
         \topsep 2\p@ plus2\p@\fi\parsep 2\p@ plus\p@ minus\p@%
         \itemsep \parsep\leftmargin\labelwidth\advance\leftmargin#2\relax}

\def\ex{\@ifnextchar [{\@ex}{\item}}
\def\@ex[#1]#2{\item\@exj[#1]{#2}}
\def\@exj[#1]#2{\@exjbg{#1} #2 \end{list}}
\def\exi#1{\item[#1]\@ifnextchar [{\@exj}{}}
\def\judgewidth#1{\def\@jwidth{#1}}
\judgewidth{??}
\def\@exjbg#1{\begin{list}{#1}{\@subex{\@jwidth}{.5ex}}\item}

\newlength{\lcommentsep}
\long\def\lcomment#1%
   {\vspace{\lcommentsep}
    \item[]\hspace*{-\leftmargin}%
    \@tempskipa=\linewidth%
    \addtolength{\@tempskipa}{\rightmargin}%
    \addtolength{\@tempskipa}{\leftmargin}%
    \parbox{\@tempskipa}{#1}%
    \vspace{\lcommentsep}%
   }

\def\attop#1{\leavevmode\vtop{\strut\vskip-\baselineskip\vbox{#1}}}

\def\pointerup{\hbox to 0pt{\hss
  \vbox{\offinterlineskip\vskip-1pt\hbox{\elevenex\char'170}\null}\hss}}
\def\pointerdown{\hbox to 0pt{\hss
  \vtop{\offinterlineskip\null\hbox{\elevenex\char'171}\vskip-1pt}\hss}}

\font\elevenex=cmex10 scaled\magstephalf  

\def\bb#1{\ifmmode\overline{\mathrm{#1}}\else$\bar{\mathrm{#1}}$\fi}

\def\bovenop#1#2{\raisebox{-0.06ex}[0ex][0ex]{$\stackrel{#1}{\mathrm{#2}}$}}
\def\vl{\rule{0.05em}{0.30em}}
\def\|#1{\ifmmode\vert#1\else\bovenop{\vl}{#1}\fi}

  \makeatother

  \makeatletter

\let\@gsingle=1
\def\singlegloss{\let\@gsingle=1}
\def\nosinglegloss{\let\@gsingle=0}
\@ifundefined{new@fontshape}%
   {\def\@selfnt{\ifx\@currsize\normalsize\@normalsize\else\@currsize\fi}}
   {\def\@selfnt{\selectfont}}

\def\gll
   {\begin{flushleft}
     \ifx\@gsingle1
        \vskip\baselineskip\def\baselinestretch{1}%
        \@selfnt\vskip-\baselineskip\fi%
    \bgroup
    \twosent
   }

\def\glll
   {\begin{flushleft}
     \ifx\@gsingle1
        \vskip\baselineskip\def\baselinestretch{1}%
        \@selfnt\vskip-\baselineskip\fi%
    \bgroup
    \threesent
   }

\def\glt{\vskip.17\baselineskip}

\newbox\lineone
\newbox\linetwo%
\newbox\linethree%
\newbox\wordone
\newbox\wordtwo%
\newbox\wordthree%
\newbox\gline
\newskip\glossglue
\newif\ifnotdone

\@ifundefined{eachwordone}{\let\eachwordone=\rm}{\relax}
\@ifundefined{eachwordtwo}{\let\eachwordtwo=\rm}{\relax}
\@ifundefined{eachwordthree}{\let\eachwordthree=\rm}{\relax}

\def\lastword#1#2#3
   {\setbox#2=\vbox{\unvbox#2%
                    \global\setbox#3=\lastbox%
                   }%
    \ifvoid#3\global\setbox#3=\hbox{#1\strut{} }\fi
   }

\def\testdone
   {\ifdim\ht\lineone=0pt
         \ifdim\ht\linetwo=0pt \notdonefalse 
         \else\notdonetrue
         \fi
    \else\notdonetrue
    \fi
   }

\gdef\getwords(#1,#2)#3 #4\\
   {\setbox#1=\vbox{\hbox{#2\strut#3 }
                    \unvbox#1%
                   }%
    \def\more{#4}%
    \ifx\more\empty\let\more=\donewords
    \else\let\more=\getwords
    \fi
    \more(#1,#2)#4\\%
   }

\gdef\donewords(#1,#2)\\{}%

\gdef\twosent#1\\ #2\\{
    \getwords(\lineone,\eachwordone)#1 \\%
    \getwords(\linetwo,\eachwordtwo)#2 \\%
    \loop\lastword{\eachwordone}{\lineone}{\wordone}%
         \lastword{\eachwordtwo}{\linetwo}{\wordtwo}%
         \global\setbox\gline=\hbox{\unhbox\gline
                                    \hskip\glossglue
                                    \vtop{\box\wordone   
                                          \nointerlineskip
                                          \box\wordtwo
                                         }%
                                   }%
         \testdone
         \ifnotdone
    \repeat
    \egroup 
   \gl@stop}

\gdef\threesent#1\\ #2\\ #3\\{
    \getwords(\lineone,\eachwordone)#1 \\%
    \getwords(\linetwo,\eachwordtwo)#2 \\%
    \getwords(\linethree,\eachwordthree)#3 \\%
    \loop\lastword{\eachwordone}{\lineone}{\wordone}%
         \lastword{\eachwordtwo}{\linetwo}{\wordtwo}%
         \lastword{\eachwordthree}{\linethree}{\wordthree}%
         \global\setbox\gline=\hbox{\unhbox\gline
                                    \hskip\glossglue
                                    \vtop{\box\wordone   
                                          \nointerlineskip
                                          \box\wordtwo
                                          \nointerlineskip
                                          \box\wordthree
                                         }%
                                   }%
         \testdone
         \ifnotdone
    \repeat
    \egroup 
   \gl@stop}

\def\gl@stop{{\hskip -\glossglue}\unhbox\gline\end{flushleft}}

 \makeatother

  \makeatletter
\begingroup
\def\x#1#2#3#4#5#6#7\relax{\def\x{#1#2#3#4#5#6}}
\expandafter\x\fmtname xxxxxx\relax \def\y{splain}
\ifx\x\y   
\gdef\SetFigFont#1#2#3{%
  \ifnum #1<17 \tiny\else \ifnum #1<20 \small\else
  \ifnum #1<24 \normalsize\else \ifnum #1<29 \large\else
  \ifnum #1<34 \Large\else \ifnum #1<41 \LARGE\else
     \huge\fi\fi\fi\fi\fi\fi
  \csname #3\endcsname}
\else
\gdef\SetFigFont#1#2#3{\begingroup
  \count@#1\relax \ifnum 25<\count@ \count@25 \fi
  \def\x{\endgroup\@setsize\SetFigFont{#2pt}}%
  \expandafter\x
    \csname \romannumeral\the\count@ pt\expandafter\endcsname
    \csname @\romannumeral\the\count@ pt\endcsname
  \csname #3\endcsname}
\fi
\endgroup

  \makeatother

\newtheorem{theorem}{Theorem}

\newcommand{\nfrac}[2]{$\frac{\mbox{#1}}{\mbox{#2}}$}

\newcommand{\F}{\mbox{${\cal F}$}}
\newcommand{\I}{{\cal I}}
\newcommand{\PP}{{\cal P}}
\newcommand{\R}{{\cal R}}

\newcommand{\U}{\mbox{${\cal U}$}}
\newcommand{\V}{{\cal V}}

\newcommand{\avm}[1]{\mbox{\scriptsize \(
                             \setlength{\arraycolsep}{.4ex}
                             \renewcommand{\arraystretch}{1.0}
                             \hspace*{-1.4ex}\left[
                             \begin{array}{ll}
                             \\[-1ex] #1 \\[-1ex]
                             \end{array}
                             \right]\hspace*{-1.7ex}
                           \)
                    }}

\newcommand{\attval}[2]{\hspace*{-1mm}
                        \mbox{\uppercase{#1}}
                        &
                        #2 \hspace*{-.25ex}\\}

\newcommand{\ind}[1]{\mbox{${\setlength{\fboxsep}{0.5mm}
        \fbox{{\tiny #1}} }$}}

\exewidth{?}
\judgewidth{?}

\author{
Suresh Manandhar \\
Language Technology Group \\
Human Communication Research Centre \\
University of Edinburgh, Scotland\\
{\tt email: Suresh.Manandhar@ed.ac.uk}
}

\title{Deterministic Consistency Checking of LP Constraints}

\begin{document}
\maketitle

\begin{abstract}
 We provide a constraint based computational model of linear
  precedence as employed in the HPSG grammar formalism. An extended
  feature logic which adds a wide range of constraints involving
  precedence is described. A sound, complete and terminating
  deterministic constraint solving procedure is given.  Deterministic
  computational model is achieved by weakening the logic such that it is
  sufficient for linguistic applications involving word-order.
\end{abstract}

{\bf Subject areas:} feature logic, constraint based grammars

\section{Introduction}
Within HPSG \cite{Pollard:hpsg1} \cite{Pollard:hpsg2} the {\em
  constituent ordering principle} given in (\ref{ex:COP1}) is intended to
express the relation between the value of the {\sc phon} attribute and
the {\sc dtrs} attribute which contains the hierarchical structure of
the derivation.
\begin{exe}
\item                                  \label{ex:COP1}
$phrasal\_sign =
 \avm{\attval{phon}{order\_constituent(\ind{1})}
      \attval{dtrs}{\ind{1}}}$
\item                                   \label{ex:LP1}
     Linear Precedence Constraint 1 (LP1):\\
      \hspace*{5ex}$HEAD[LEX +]\ <\ []$
\end{exe}
However, it is not entirely clear how {\em order\_constituent} is
supposed to interpret various linear precedence statements such as LP1.

\subsection{Reape's approach}
The idea taken in Reape's approach \cite{Reape:ThingsInOrder} is to
suggest that word-order is enforced between locally definable word order
domains which are ordered sequences of constituents. Word order domains
in Reape's approach are totally ordered sequences. A domain union
operation as given in (\ref{ex:DomainUnion}) is then employed to
construct word order domains locally within a HPSG derivation step.
\begin{exe}
\item                                 \label{ex:DomainUnion}
$\begin{array}[t]{l}
  \bigcirc(\eta, \eta, \eta).\\
\bigcirc(x \circ \sigma_{1}, \sigma_{2}, x \circ \sigma_{3}) \leftrightarrow
  \bigcirc(\sigma_{1}, \sigma_{2}, \sigma_{3}).\\
\bigcirc(\sigma_{1}, x \circ \sigma_{2}, x \circ \sigma_{3}) \leftrightarrow
  \bigcirc(\sigma_{1}, \sigma_{2}, \sigma_{3}).
 \end{array}$
\end{exe}
If $A$ is the string $<a, b>$ and $B$ is the string \mbox{$<c,d>$}, their
domain union $C$ given by $\bigcirc(A, B, C)$ will produce all the
sequences in which $a$ precedes $b$ and $c$ precedes $d$ {\em i.e.} the
following sequences:
\vspace*{-1ex}
\begin{itemize}
\item[] $\begin{array}{lll}
<a, b, c, d> & <a, c, b, d> \\
<a, c, d, b> & <c, d, a, b> \\
<c, d, a, b> & <c, a, b, d>
\end{array}$
\end{itemize}
\vspace*{-1ex}
However in this system to encode the property that $\{x, y, z\}$ is a domain
in which the ordering is arbitrary ({\em i.e.} free) then one needs the
following disjunctive statements:
\vspace*{-1ex}
\begin{itemize}
\item[] $\begin{array}{lllll}
         <x, y, z> \sqcup <x, z, y> \sqcup\\
         <y, x, z> \sqcup <y, z, x> \sqcup\\
         <z, x, y> \sqcup <z, y, x>
         \end{array}$
\end{itemize}
\vspace*{-1ex}
It is simply not possible to be agnostic about the relative ordering of
sequence elements within Reape's system.

We identify two deficiencies in Reape's approach namely:
\vspace*{-1ex}
\begin{itemize}
\item System is non-deterministic (generate and test paradigm)

\item Not possible to be agnostic about order
\end{itemize}
This is so since domain union is a {\em non-deterministic} operation and
secondly underspecification of ordering within elements of a domain is
not permitted.

In the following sections we describe a constraint language for
specifying LP constraints that overcomes both these deficiencies.
Additionally our constraint language provides a broad range of
constraints for specifying linear precedence that go well beyond what is
available within current typed feature formalisms. Our approach is in
the spirit of Reape's approach but improves upon it.

Furthermore, a sound, complete and terminating consistency checking
procedure is described. Our constraint solving rules are {\em
  deterministic} and {\em incremental}. Hence these do not introduce
costly choice-points. These constraint solving rules can be employed for
building an efficient implementation. This is an important requirement
for practical systems.  Indeed we have successfully extended the ProFIT
typed feature formalism \cite{Erbach:ProFIT} with the constructs
described in this paper.

\section{Outline of an alternative approach}
To motivate our approach we start with an example on scrambling in German
subordinate clauses.
\vspace*{-3ex}
{\judgewidth{}
\begin{exe}
\item
\gll da{\ss}  er      einen Mann      in der Stra{\ss}e laufen sah.\\
     that     he      a     man      in the street     walking saw.\\
\glt that he saw a man walking in the street.
\item da{\ss} er  in der Stra{\ss}e  einen Mann  laufen  sah.
\item da{\ss} einen Mann  er  in der Stra{\ss}e  laufen  sah.
\item da{\ss} einen Mann  in der Stra{\ss}e  er  laufen  sah.
\item da{\ss} in der Stra{\ss}e  er  einen Mann  laufen  sah.
\item da{\ss} in der Stra{\ss}e  einen Mann  er  laufen  sah.
\label{ex:GermanEx2L}
\end{exe}}
\vspace*{-1ex}
The above data can be captured precisely if we can state
that {\em sah} requires both its verbal argument {\em laufen} and its NP
argument {\em er} to precede it. Similarly, {\em laufen} would require
both its arguments {\em einen Mann} and {\em in der Stra{\ss}e} to
precede it. This is illustrated schematically in (\ref{ex:germansub})
below.
\begin{exe}
\item                          \label{ex:germansub}
\hspace*{2ex}\attop{
\vspace*{-2.8ex}\epsffile{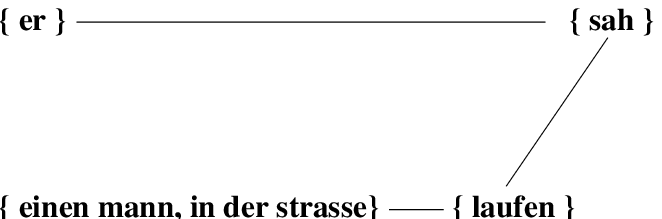}
}
\end{exe}
Our idea is to employ a specification such as the one given in
(\ref{ex:German}) which is a partial specification of the lexical entry
for the verb {\em sah}. The specification can be thought of as a
formal specification of the intuitive description given in
(\ref{ex:germansubi}).
\begin{exe}
\item                                              \label{ex:German}
\hspace*{-2ex}{\small
\attop{
$\left(
\hspace*{-2ex}\begin{array}{l}
\begin{array}{llll}
 V\ \sqcap & phon:& \!\!\! <sah>\ \sqcap \\
          & field:& \!\!\! Field\ \sqcap\\
          & syn: ( & \!\!\! cat:v\ \sqcap \\
          &        & \!\!\! subcat: \{ \begin{array}[t]{ll}
                                NP\ \sqcap dom:NPdom,\\
                                V_{i}\ \sqcap dom:V_{i}dom\}\ \sqcap
                                \end{array}\\
          &        & \!\!\! dom: \supseteq NPdom\ \sqcap \\
          &        & \!\!\! dom: \supseteq V_{i}dom\ )\ \sqcap
\end{array}
\\
\begin{array}{ll}
                &  V_{i}dom <_{dom}\ \ \{V\}\ \sqcap\\
                &  NPdom <_{dom}\ \ \{V_{i}\}\ \sqcap\\
                &  V_{i} < V
\end{array}
\end{array}
\hspace*{-4ex}\right)$
}
}
\end{exe}
For space reasons, our treatment is necessarily somewhat superficial
since we do not take into account other interacting phenomena such as
{\em fronting} or {\em extraposition}.

The definition in   (\ref{ex:German}) does not  make specific  assumption
about  whether a context-free backbone is  employed or not. However, if a
CFG backbone  is  employed then we   assume that the   value of  the {\em
  subcat} attribute is treated as an {\em unordered} sequence ({\em i.e.}
a set) as defined in (\ref{ex:German}).
\begin{exe}
\item                          \label{ex:germansubi}
\attop{
\vspace*{-3ex}
\hspace*{2ex}\setlength{\unitlength}{0.0063in}
\begin{picture}(397,146)(0,-10)
\path(295,20)(385,115)
\path(80,125)(370,125)
\path(65,5)(245,5)
\put(0,120){\makebox(0,0)[lb]{\smash{{{\SetFigFont{8}{9.6}{bf}$NPdom$}}}}}
\put(0,0){\makebox(0,0)[lb]{\smash{{{\SetFigFont{8}{9.6}{bf}$V_{i}dom$}}}}}
\put(265,0){\makebox(0,0)[lb]{\smash{{{\SetFigFont{8}{9.6}{bf}$V_{i}$}}}}}
\put(385,120){\makebox(0,0)[lb]{\smash{{{\SetFigFont{8}{9.6}{bf}$V$}}}}}
\end{picture}
}
\end{exe}
The essential idea is to use set-valued descriptions to model word-order
domains. In particular subset constraints \cite{Manandhar:Sets} are
employed to construct larger domains from smaller ones. Thus in example
(\ref{ex:German}) the domain of the verb is constructed by including the
domains of the subcategorised arguments (enforced by the constraints
$dom: \supseteq NPdom \sqcap dom: \supseteq V_{i}Dom$). Note that in this
example the verb itself is not part of its own domain. The binary
constraint $V_{i} < V$ enforces precedence ordering between the {\em
  signs} $V_{i}$ and $V$. The constraint $V_{i}dom <_{dom}\ \ \{V\}$
ensures that every element of the set $V_{i}Dom$ precedes the sign $V$.
In other words, the set $V_{i}Dom$ is in the {\em domain precedence}
relation with the singleton $\{V\}$.

However there are strong  constraints on ordering in the middle
field. For instance, when pronomial complements are involved then not all
permutations are acceptable. Examples such as (\ref{ex:UnGerman}) are
considered ungrammatical.
\begin{exe}
\item *da{\ss} in der Stra{\ss}e  ihn  er  laufen  sah. \label{ex:UnGerman}
\end{exe}
According to Uszkoreit \cite{Uszkoreit:ConstraintsOnOrder}, ordering of
arguments in the middle field is governed by the following set of LP
constraints given in (\ref{ex:LPrules}) which are to be interpreted
disjunctively.
\vspace*{-.5ex}
\begin{exe}
\item                           \label{ex:LPrules}
{\small $\begin{array}[t]{l}
        PPRN:+\ <\ PPRN:-\\
        TR:agent\ <\ TR:theme\\
        TR:agent\ <\ TR:goal\\
        TR:goal\ <\ TR:theme\\
        FOCUS:-\ <\ FOCUS:+
       \end{array}$}
\end{exe}
\vspace*{-.5ex}
The LP constraint in (\ref{ex:LPrules}) states that for every pair of
constituents in the middle field at least one of the conditions should
apply otherwise the sentence is considered ungrammatical. A related but
more elaborate LP rule mechanism is considered in
\cite{Steinberger:TreatingWOinMT}.

To approximate this complex LP constraint employing the kind of logical
machinery described in this paper, we can use a description such as the
one given in  (\ref{ex:LPencoding}). The definition given in
(\ref{ex:LPencoding}) extends the description given in (\ref{ex:German}).
\begin{exe}
\item                           \label{ex:LPencoding}
{\small  \hspace*{-2ex}$\begin{array}[t]{l}
       syn:dom: MF\ \sqcap\\
      \exists x \exists y\ if\ x \in MF \wedge y \in MF \wedge x < y\\
      \ \hspace*{5ex}then\\
      \ \hspace*{8ex}\begin{array}{ll}
                       if & \hspace*{-3ex} x = pprn:+ \wedge y = pprn:- \\
                          & \hspace*{-3ex} then\ \top\\
                       else\\
                       if & \hspace*{-3ex} x = tr:agent \wedge y = tr:theme \\
                          & \hspace*{-3ex} then\ \top\\
                       else\\
                       if & \hspace*{-3ex} x = tr:agent \wedge y = tr:goal \\
                          & \hspace*{-3ex} then\ \top\\
                       else\\
                       if & \hspace*{-3ex} x = tr:goal \wedge y = tr:theme \\
                          & \hspace*{-3ex} then\ \top\\
                       else\\
                          & \hspace*{-3ex} x = focus:- \wedge y = focus:+
                     \end{array}
     \end{array}$
}
\end{exe}
The definition in (\ref{ex:LPencoding}) can be understood as follows.
The feature constraint $syn:dom: MF$ co-instantiates the middle field
domain to the variable $MF$. To keep the example simple, we assume that
the whole domain is in the middle field and we ignore {\em fronting} or
{\em extraposition}.  A more complex condition would be needed to handle
these.

The rest of the definition in (\ref{ex:LPencoding}) ensures that for
every pair of elements $x$ and $y$ such that $x$ and $y$ are both members
of $MF$ and $x$ {\em precedes} $y$ at least one of the LP constraints
hold. If every LP constraint is violated then an inconsistency results.
The constraints in (\ref{ex:LPencoding}) is a weaker representation of
the disjunctive specification given in (\ref{ex:LPencoding2}).
\begin{exe}
\item                               \label{ex:LPencoding2}
{\small
\hspace*{-1ex}$\exists x \exists y\ if\ (x \in MF \wedge y \in MF \wedge x <
y)$\\
          \hspace*{10ex}$then$\\
          \hspace*{10ex}$\bigvee\left\{\begin{array}{l}
                       x = pprn:+ \wedge y = pprn:- \\
                       x = tr:agent \wedge y = tr:theme \\
                       x = tr:agent \wedge y = tr:goal \\
                       x = tr:goal \wedge y = tr:theme \\
                       x = focus:- \wedge y = focus:+
                     \end{array}\right\}$
}
\end{exe}
The description in (\ref{ex:LPencoding2}) non-deterministically requires
that at least one of the LP constraints hold. On the other hand, the
description in (\ref{ex:LPencoding}) {\em waits} until either one of the
LP constraints is satisfied (in which case it succeeds) or all the LP
constraints are violated (in which case it fails). Thus the description
in (\ref{ex:LPencoding}) can be solved deterministically.

Thus (\ref{ex:LPencoding}) should rule out the ungrammatical example in
(\ref{ex:UnGerman}) if the assumptions regarding {\em focus} are made as
in (\ref{ex:UnGerman2}).
\hspace*{-1ex}\begin{exe}
\item  \label{ex:UnGerman2}
  \hspace*{-2ex}*da{\ss} \begin{tabular}[t]{l}
             \underline{in der Stra{\ss}e}\\
              {\em pprn}:-\\
              {\em th}:{\em theme}
            \end{tabular}\hspace*{-2ex}
          \begin{tabular}[t]{l}
              ihn\\
              {\em focus}:-\\
              {\em pprn}:+\\
              {\em tr}:{\em agent}
          \end{tabular}
  \hspace*{-2ex}er  laufen sah.
\end{exe}
Note that it is not necessary to know whether the PP {\em in der
  Stra{\ss}e} is focussed to rule out (\ref{ex:UnGerman2}) since the fact
that the pronoun {\em ihn} is \mbox{{\em focus}:-} is enough to trigger
the inconsistency.

\section{Some generic LP constraints}
As suggested by the example in (\ref{ex:German}), in general we would
want support within typed feature formalisms for at least the following
kinds of LP constraints.
\begin{enumerate}
\item Sign$_{1}  <$ Sign$_{2}$

\item Dom$_{1}  <_{dom}$ Dom$_{2}$\\
(Dom$_{1}$ and Dom$_{2}$ are set-valued)

\item Dom$_{1}$ {\em is included in} Dom$_{2}$
\end{enumerate}
The constraint Sign$_{1} <$ Sign$_{2}$ states that Sign$_{1}$ {\em
  precedes} Sign$_{2}$. The constraint Dom$_{1} <_{dom}$ Dom$_{2}$ states
that every element of the set described by Dom$_{1}$ precedes every
element of the set described by Dom$_{2}$.  Constraints such as Dom$_{1}$
{\em is included in} Dom$_{2}$ essentially builds larger domains from
smaller ones and can be thought of as achieving the same effect as
Reape's domain union operation. Note crucially that within our approach
the specification of precedence constraints (such as Sign$_{1} <$
Sign$_{2}$ and Dom$_{1} <_{dom}$ Dom$_{2}$) is independent of the domain
building constraint ({\em i.e.} the constraint Dom$_{1}$ {\em is included
  in} Dom$_{2}$). This we believe is a generalisation of Reape's
approach.

Other constraints such as the following involving {\em immediate
  precedence} and {\em first element of a domain} are of lesser
importance. However, these could be of the form:
\vspace*{-1ex}
\begin{enumerate}

\item Sign$_{1}$ {\em immediately-precedes} Sign$_{2}$

\item {\em First daughter of} Dom$_{1}$  is Sign$_{1}$

\end{enumerate}
\vspace*{-1ex} To be able to state descriptions such as in
(\ref{ex:LPencoding}), we also want to introduce {\em guarded} (or {\em
  conditional}) LP constraints such the following:
\begin{enumerate}
\item {\em if} Sign$_{1}$ is  NP[acc] $\wedge$  Sign$_{2}$
      is NP[dat]\\
     \ \hspace*{3ex} {\em then} Sign$_{1}$ $<$ Sign$_{2}$\\
( Guards on Feature constraints)

\item {\em if} Sign$_{1}  <$ Sign$_{2}$ {\em then} $\ldots \ldots$\\
( Guards on precedence constraints)

\item $\exists x \exists y$ ({\em if} x:NP[acc] $\in$ Dom $\wedge$ \\
 \ \hspace*{3ex}y:NP[dat] $\in$ Dom \\
     {\em then} x $<$ y)\\
( Guards on set members)
\end{enumerate}
Guarded constraints can be thought of as {\em conditional constraints}
whose execution depends on the presence of other constraints. The
condition part $G$ of a guarded constraint $if\ G\ then\ S\ else\ T$ is
known as a {\em guard}. The consequent $S$ is executed if the current set
of constraints {\em entail} the guard $G$. The consequent $T$ is executed
if the current set of constraints {\em disentail} the guard $G$.  If the
current set of constraints neither entail nor disentail $G$ then the
execution of the whole guarded constraint is {\em blocked} until more
information is available.

The application of guarded constraints within computational linguistics
has not been well explored. However, the {\em Horn extended feature
  structures} described in \cite{Hegner:Horn} can be thought of as adding
guards to feature structures. On the other hand, within logic programming
guarded logic programming languages have a longer history originating
with {\em committed-choice languages} \cite{Ueda:GHC} and popularised by
the {\em concurrent constraint programming} paradigm due to {\em
  Saraswat} \cite{Saraswat:CCP} \cite{Saraswat:CCPBook}.

For space reasons, we do not cover the logic of guarded feature
constraints, guards on set membership constraints and guards on
precedence constraints. Guarded feature constraints have been extensively
studied in \cite{AitKaci:constraint} \cite{Treinen:records}
\cite{AitKaciPodelski:Functions}.
\section{A feature logic with LP constraints}
In this section we provide formal definitions for the syntax and
semantics of an extended feature logic that directly supports linear
precedence constraints as logical primitives. The logic described in this
paper is a further development of the one described in
\cite{Manandhar:Thesis}.

The syntax of the constraint language is defined by the following BNF
definitions.

\vspace{0.5ex}
{\large \bf Syntax}
\vspace{0.5ex}

Let $\F$ be the set of relation symbols and let $\PP$ be the set of
irreflexive relation symbols. We shall require that $\F$ and $\PP$ are
disjoint.
\vspace*{-2ex}
\begin{center}
$\begin{array}{lll}
\phi, \psi \longrightarrow
 & x =  f:y                & \mbox{\rm feature constraint}\\
 & x = \exists f:y          & \mbox{\rm set-membership}\\
 & x = \exists p^{+}:y      & \mbox{\rm transitive closure}\\
 & x = \exists p^{*}:y      & \mbox{\rm reflex-trans closure}\\
 & x = f: \supseteq g(y)    & \mbox{\rm subset inclusion}\\
 & x = [f\ p\ 1] y          & \mbox{\rm first daughter}\\
 & f(x):p^{+}:g(y)          & \mbox{\rm domain precedence}\\
 & f(x):p^{*}:g(y)          & \mbox{\rm domain prec. equals}\\
 & \phi\ \&\ \psi           & \mbox{\rm conjunction}\\
\multicolumn{3}{l}{\mbox{where $f \in \F$ and $p \in \PP$}}
\end{array}$
\end{center}
The constraint $x = f: y$ specifies that $y$ is the {\em only} $f$-value
of $x$. The constraint $x = \exists f: y$ states that $y$ is one of the
$f$-values of $x$.

The constraint $x = \exists p^{+}: y$ just says that $x$ is related to
$y$ via the transitive closure of $p$. The precedence constraint such as
Sign$_{1}$ {\em precedes} Sign$_{2}$ is intended to be captured by
the constraint Sign$_{1} = \exists p^{+}: $Sign$_{2}$ where $p$ denotes
the (user chosen) {\em immediate precedence} relation.

Similarly, $x = \exists p^{*}: y$ states that $x$ is related to $y$ via
the transitive, reflexive closure of $p$. This constraint is similar to
the constraint $x = \exists p^{+}: y$ except that it permits $x$ and $y$
to be equal.

The constraints $f(x):p^{+}:g(y)$ and $f(x):p^{*}:g(y)$ are intended to
enforce precedence between two word-ordering domains. The constraint
$f(x):p^{+}:g(y)$ states that every $f$-value of $x$ {\em precedes} ({\em
  i.e.} is in the $p^{+}$ relation with) every $g$-value of $y$. The
constraint $f(x):p^{*}:g(y)$ is analogous.

The constraint $x = [f\ p\ 1] y$ states that $y$ is the {\em first
  daughter} amongst the $f$-values of $x$ ({\em i.e.} is in the $p^{*}$
relation with every $f$-value of $x$).

Since our language supports both feature constraints and set-membership
constraints the conventional semantics for feature logic
\cite{Smolka:constraint} needs to be extended. The essential difference
being that we interpret every feature/relation as a binary relation on
the domain of interpretation. Feature constraints then require that they
behave functionally on the variable upon which the constraint is
expressed.

A precise semantics of our constraint language is given next.

{\vspace{0.5ex}
\large \bf Semantics}
\vspace{0.5ex}

An interpretation structure $\I = < \U^{I}, .^{I}>$ is a structure such
that:
\vspace*{-0.5ex}
\begin{itemize}
\item $\U^{I}$ is an arbitrary non-empty set
\item $.^{I}$ is an interpretation function which maps:
  \begin{itemize}
  \item every relation $f \in \F$ to a binary relation: $f^{I} \subseteq
    \U^{I} \times \U^{I}$
  \item every relation $p \in \PP$ to a binary relation: $p^{I} \subseteq
    \U^{I} \times \U^{I}$ with the added condition that
     $(p^{I})^{+}$ is irreflexive
  \end{itemize}
\end{itemize}
\vspace*{-0.5ex}
A variable assignment $\alpha$ is a function \mbox{$\alpha: \V \longrightarrow
\U^{I}$}.

\vspace*{0.5ex}
We shall write $f^{I}(e)$ to mean the set:
\vspace*{-0.5ex}
\begin{itemize}
\item[] $f^{I}(e) = \{ e' \in \U^{I} \mid (e,e') \in f^{I} \}$
\end{itemize}
\vspace*{-0.5ex}
We say that an interpretation $\I$ and a variable assignment $\alpha$
satisfies a constraint $\phi$ written $\I, \alpha \models \phi$ if the
following conditions are satisfied:

\arraycolsep = .4\arraycolsep
\hspace*{-3ex}$\begin{array}{lll}
 \I, \alpha \models \phi\ \&\ \psi & \Longleftrightarrow &
 \I, \alpha \models \phi \wedge   \I, \alpha \models \psi \\
 \I, \alpha \models x =  f:y & \Longleftrightarrow &
       f^{I}(\alpha(x)) = \{\alpha(y)\}\\
 \I, \alpha \models x = \exists f:y  & \Longleftrightarrow &
 (\alpha(x),\alpha(y)) \in f^{I}\\
 \I, \alpha \models x = \exists p^{+}:y & \Longleftrightarrow &
 (\alpha(x),\alpha(y)) \in (p^{I})^{+}\\
 \I, \alpha \models x = \exists p^{*}:y & \Longleftrightarrow &
 (\alpha(x),\alpha(y)) \in (p^{I})^{*}\\
 \I, \alpha \models x = f: \supseteq g(y) & \Longleftrightarrow &
 f^{I}(\alpha(x)) \supseteq g^{I}(\alpha(y))
\end{array}$

\hspace*{-3ex}$\begin{array}{lll}
 \I, \alpha \models x = [f\ p\ 1] y  & \Longleftrightarrow &
       \alpha(y) \in f^{I}(\alpha(x)) \wedge\\
 &  &  \forall e \in \U^{I}\\
 &  & (e \in f^{I}(\alpha(x)) \Rightarrow\\
 &  & (\alpha(y),e) \in (p^{I})^{*})\\
 \I, \alpha \models f(x):p^{+}:g(y)  & \Longleftrightarrow &
\forall e_{1}, e_{2} \in \U^{I}\\
 &  & ((e_{1} \in f^{I}(\alpha(x)) \wedge\\
 & &  \ e_{2} \in g^{I}(\alpha(y)))\\
 &  &  \Rightarrow
                              (e_{1},e_{2}) \in (p^{I})^{+})\\
\I, \alpha \models f(x):p^{*}:g(y) & \Longleftrightarrow &
\forall e_{1}, e_{2} \in \U^{I}\\
 &  & ((e_{1} \in f^{I}(\alpha(x)) \wedge\\
 &  & \  e_{2} \in g^{I}(\alpha(y)))\\
 &  &  \Rightarrow
                              (e_{1},e_{2}) \in (p^{I})^{*})
\end{array}$
\vspace{0.5ex}
Given the above semantics, it turns out that the {\em first-daughter}
constraint can be defined in terms of other constraints in the
logic. Let $f\_p\_1$ be a distinct relation symbol then we can
equivalently define the first-daughter constraint by:
\begin{itemize}
\item $\begin{array}[t]{l}
       x = [f\ p\ 1] y \approx
       x = f\_p\_1: y \wedge \\
       x = \exists f:y \wedge
       f\_p\_1(x):p^{*}:f(x)
      \end{array}$
\end{itemize}
The translation states that $y$ (which is the $f\_p\_1$-value of $x$)
precedes or is equal to every $f$-value of $x$ and $y$ is a $f$-value of
$x$. For this to work, we require that the feature symbol $f\_p\_1$
appears only in the translation of the constraint $x = [f\ p\ 1] y$.

\subsection{Two Restrictions}
The logic we have described comes with 2 limitations which at first
glance appears to be somewhat severe, namely:
\vspace*{-1ex}
\begin{itemize}
\item {\bf NO} {\em atomic values}
\item {\bf NO} {\em precedence as a feature}
\end{itemize}
This is so because it turns out that adding both functional precedence
and atoms in general leads to a non-deterministic constraint solving
procedure. To illustrate this, consider the following constraints:
\vspace*{-1ex}
\begin{itemize}
\item[] $x =  f:y \wedge y = a \wedge x = \exists f^{*}:z$
\end{itemize}
\vspace*{-1ex}
where $a$ is assumed to be an {\em atom}.

The above constraints state that $y$ is the $f$-value of $x$ and $y$ is
the atom $a$ and $z$ is related to $x$ by the reflexive-transitive
closure of $f$.

Determining consistency of such constraints in general involves
solving for the following disjunctive choices of constraints.
\vspace*{-1ex}
\begin{itemize}
\item[] $x = z$ or $y = z$
\end{itemize}
\vspace*{-1ex}
However for practical reasons we want to eliminate any form of
backtracking since this is very likely to be expensive for implemented
systems. On the other hand, we certainly cannot prohibit atoms since they
are crucially required in grammar specification. But disallowing
functional precedence is less problematic from a grammar development
perspective.

\subsection{Imposing the restriction}
We note that precedence can be restricted to non-atomic types such as
HPSG {\em signs} without compromising the grammar in any way. We then
need to ensure that precedence constraints never have to consider atoms
as their values. This can be easily achieved within current typed feature
formalisms by employing {\em appropriateness conditions}
\cite{Carpenter:typed}.

An {\em appropriateness condition} just states that a given feature (in
our case a relation) can only be defined on certain (appropriate) types.
The assumption we make is that precedence is specified in such a way that
is  appropriate only for non-atomic types. This restriction can be
imposed by the system ({\em i.e.} a typed feature formalism) itself.

\section{Constraint Solving}
We are now ready to consider consistency checking rules for our
constraint language. To simplify the presentation we have split up the
rules into two groups given in figure \ref{fig:ConstraintSolving1} and
figure \ref{fig:ConstraintSolving2}.

\begin{figure}
\begin{center}
\begin{list}{}{\leftmargin 10ex
\labelwidth 10ex}

\item[(Equals)] \nfrac{$x = y \wedge C_{s}$}{$
              x = y \wedge [x/y]C_{s}$}\\
      if $x \neq y$ and $x$ occurs in $C_{s}$

\item[(Feat)] \nfrac{$x = f: y \wedge x = f: z \wedge C_{s}$}{$
             x = f: y \wedge y = z \wedge C_{s}$}

\item[(FeatExists)] \nfrac{$x = f: y \wedge x = \exists f: z \wedge C_{s}$}{$
             x = f: y \wedge x = \exists f: z \wedge
             y = z \wedge C_{s}$}

\item[(Subset)]
 \hspace*{-4ex}\nfrac{$x = f : \supseteq g(y) \wedge y = G: z \wedge C_{s}$}{$
       x = \exists f: y \wedge x = f : \supseteq g(y) \wedge
       y = G: z \wedge C_{s}$}\\
    if $x = \exists f: y \not\in C_{s}$\\
    where $G$ ranges over $g, \exists g$




\end{list}
\end{center}
\caption{Constraint Solving - I}
\label{fig:ConstraintSolving1}
\end{figure}

The constraint solving rules given in figure \ref{fig:ConstraintSolving1}
deal with constraints involving {\em features}, {\em set-memberships},
{\em subset} and {\em first daughter}. Rules (Equals) and (Feat) are the
usual feature logic rules \cite{Smolka:constraint} that deal with
equality and features. By $[x/y]C_{s}$ we mean replacing every occurrence
of $x$ with $y$ in $C_{s}$. Rule (FeatExists) deals with the interaction
of feature and set-membership constraint.  Rule (Subset) deals with
subset constraints and adds a new constraint $x = \exists f: y$ in the
presence of the subset constraint $x = f : \supseteq g(y)$ and the
constraint $y = G: z$ (where $G$ ranges over $g, \exists g$).

The constraint solving rules given in figure \ref{fig:ConstraintSolving2}
deal with constraints involving the {\em precedes} and the {\em precedes
  or equal to} relations and {\em domain precedence}.  Rule (TransConj)
eliminates the weaker constraint \mbox{$x = \exists p^{*}: y$} when both
\mbox{$x = \exists p^{*}: y$} \mbox{$\wedge\ x = \exists p^{+}: y$} hold.
Rule (TransClos) effectively computes the transitive closure of the
precedence relation one-step at a time. Rule (Cycle) detects cyclic
relations that are consistent, namely, when $x$ {\em precedes or equals}
$y$ and {\em vice versa} then $x=y$ is asserted.  Finally rule (DomPrec)
propagates constraints involving domain precedence.

\begin{figure}
\begin{center}
\begin{list}{}{\leftmargin 10ex
\labelwidth 10ex}

\item[(TransConj)] \nfrac{$x = \exists p^{*}: y \wedge x = \exists p^{+}: y
      \wedge C_{s}$}{$
       x = \exists p^{+}:y \wedge C_{s}$}

\item[(TransClos)]
 \nfrac{$x = \exists R_{1}: y \wedge y = \exists R_{2}: z
      \wedge C_{s}$}{$
       \begin{array}[t]{c}
       x = \exists (R_{1}\times R_{2}): z \wedge\\
       x = \exists R_{1}: y \wedge y = \exists R_{2}: z \wedge
        C_{s}
       \end{array}$} \\
    if $x = \exists p^{+}:z \not\in C_{s} \wedge$\\
       $x = \exists (R_{1}\times R_{2}): z \not\in C_{s}$\\
  where $R_{1}\times R_{2}$ is computed from: \\
 \begin{tabular}{|c|c|c|}
  \hline
  $\times$     & $p^{*}$ & $p^{+}$ \\
  \hline
  $p^{*}$ & $p^{*}$ & $p^{+}$ \\
  $p^{+}$ & $p^{+}$ & $p^{+}$ \\
 \hline
 \end{tabular}

\item[(Cycle)] \nfrac{$x = \exists p^{*}: y \wedge y = \exists p^{*}: x
      \wedge C_{s}$}{$
       x = y \wedge C_{s}$} \\

\item[(DomPrec)]
 \nfrac{$\begin{array}{ccc}
            f(x):R:g(y) \wedge
            x = \exists f: x_{1} \wedge\\
             y = \exists g: y_{1}
            \wedge C_{s}
     \end{array}$}{$\begin{array}{ccc}
            x_{1} = \exists R: y_{1} \wedge
            f(x):R:g(y) \wedge\\
            x = \exists f: x_{1} \wedge
             y = \exists g: y_{1}
            \wedge C_{s}
       \end{array}$}\\
     if $x_{1} = \exists p^{+}:y_{1} \not\in C_{s} \wedge$\\
        \ \ $x_{1} = \exists R: y_{1} \not\in C_{s}$\\
     where $R$ ranges over $p^{+}, p^{*}$

\end{list}
\end{center}
\caption{Constraint Solving - II}
\label{fig:ConstraintSolving2}
\end{figure}

We say that a set of constraints are in {\bf normal form} if {\em no}
constraint solving rules are applicable to it. We say that a set of
constraints in normal form contains a {\bf clash} if it contains
constraints of the form:
\vspace*{-1ex}
\begin{itemize}
\item[] $x = \exists p^{+}:x$
\end{itemize}
\vspace*{-1ex}
In the following sections we show that our constraint solving rules are
sound and every {\bf clash-free} constraint system in normal form is
consistent.
\subsection{Soundness, Completeness and Termination}
\vspace*{-1ex}
\begin{theorem}[Soundness]
  Let $\I, \alpha$ be any interpretation, assignment pair and let $C_{s}$
  be any set of constraints. If a constraint solving rule transforms
  $C_{s}$ to $C'_{s}$ then:
  \begin{itemize}
  \item[] $\I, \alpha \models C_{s}$ iff $\I, \alpha \models C'_{s}$
  \end{itemize}
\end{theorem}
\vspace*{-1ex}
{\sl Proof Sketch}: The soundness claim can be verified by checking that
every rule indeed preserves the interpretation of every variable and
every relation symbol.

Let $succ(x, f)$ and $succ(x, p)$ and denote the sets:
\vspace*{-1ex}
\hspace*{-6ex}\begin{itemize}
\item $succ(x, f) =\\
       \hspace*{2ex}\{ y \mid
        x = \exists f: y \in C_{s} \vee\ x = f: y \in C_{s}
       \}$

\item $succ(x, p) =\{ y \mid\\
        \hspace*{2ex}\begin{array}[t]{l}
           x = \exists R: y \in C_{s} \wedge \\
           \neg \exists z: (x = \exists R_{1}: z \wedge
                           z = \exists R_{2}: y) \in C_{s}\}
        \end{array}$\\
        \hspace*{3ex}where $R, R_{1}, R_{2} \in \{ p^{+}, p^{*}\}$
\end{itemize}
\vspace*{-1ex}
\begin{theorem}[Completeness]
  A constraint system $C_{s}$ in normal form is consistent iff $C_{s}$ is
  clash-free.
\end{theorem}
{\em Proof Sketch}: For the first part, let $C_{s}$ be a constraint
system containing a clash then it is clear from the definition of clash
that there is no interpretation $\I$ and variable assignment $\alpha$
which satisfies $C_{s}$.

Let $C_{s}$ be a clash-free constraint system in normal form.

We shall construct~an~interpretation \mbox{$\R = <\U^{R}, .^{R}>$} and a
variable assignment $\alpha$ such that $\R, \alpha \models C_{s}$.

Let $\U^{R} = \V$.

The assignment function $\alpha$ is defined as follows:
\vspace*{-2ex}
\begin{itemize}
\item if $x$ does not occur in $C_{s}$ then $\alpha(x) = x$
\item if $x$ is such that $x$ occurs exactly once in  $x = y \in C_{s}$
 then $\alpha(x) = x$
\item if $x = y \in C_{s}$ then $\alpha(y) = \alpha(x)$
\end{itemize}
\vspace*{-1ex}
Note that for constraints in normal form: if $x = y \in C_{s}$ then
either $x$ is identical to $y$ or $x$ occurs just once in $C_{s}$ (in the
constraint $x = y$).  Otherwise Rule (Equals) is applicable.

The interpretation function $.^{R}$ is defined as follows:
\begin{itemize}
\item $f^{R}(\alpha(x)) = succ(\alpha(x), f)$
\item $p^{R}(\alpha(x)) = succ(\alpha(x), p)$
\end{itemize}

It can be shown by a case by case analysis that for every constraint $K$
in $C_{s}$:\\
$\R, \alpha \models K$.

Hence we have the theorem.

\begin{theorem}[Termination]
The~consistency checking procedure terminates in a finite number of
steps.
\end{theorem}
{\em Proof Sketch}: The termination claim can be easily verified if we
first exclude rules (Subset),
(TransClos) and (DomPrec) from
consideration. Then for the remainder of the rules termination is obvious
since these rules only simplify existing constraints. For these rules:
\begin{enumerate}
\item Rule (Subset) increases the size of $succ(x,f)$ but since none of
  our rules introduces new variables this is terminating.

\item Rules
(TransClos) and (DomPrec) asserts a relation $R$
  between pairs of variables $x, y$. However, none of these rules apply
  once $x = \exists p^{+}:y$ is known. Furthermore, if $x = \exists
  p^{+}:y$ is known it is never simplified to the weaker $x = \exists
  p^{*}:y$. This means that these rules converge.

\end{enumerate}

\begin{figure}
\begin{center}
\mbox{\setlength{\unitlength}{0.0063in}
\begin{picture}(355,315)(0,-10)
\path(170,260)(170,185)
\path(168.000,193.000)(170.000,185.000)(172.000,193.000)
\path(40,260)(40,165)(145,165)
\path(137.000,163.000)(145.000,165.000)(137.000,167.000)
\path(200,145)(200,185)(145,185)
	(145,145)(200,145)
\path(70,280)(145,280)
\path(137.000,278.000)(145.000,280.000)(137.000,282.000)
\path(200,280)(275,280)
\path(267.000,278.000)(275.000,280.000)(267.000,282.000)
\path(70,260)(70,300)(15,300)
	(15,260)(70,260)
\path(200,260)(200,300)(145,300)
	(145,260)(200,260)
\path(330,260)(330,300)(275,300)
	(275,260)(330,260)
\path(355,0)(355,40)(300,40)
	(300,0)(355,0)
\path(55,20)(100,20)
\path(92.000,18.000)(100.000,20.000)(92.000,22.000)
\path(55,0)(55,40)(0,40)
	(0,0)(55,0)
\path(155,0)(155,40)(100,40)
	(100,0)(155,0)
\path(155,20)(200,20)
\path(192.000,18.000)(200.000,20.000)(192.000,22.000)
\path(255,0)(255,40)(200,40)
	(200,0)(255,0)
\path(255,20)(300,20)
\path(292.000,18.000)(300.000,20.000)(292.000,22.000)
\put(165,160){\makebox(0,0)[lb]{\smash{{{\SetFigFont{9}{10.8}{rm}C}}}}}
\put(35,275){\makebox(0,0)[lb]{\smash{{{\SetFigFont{9}{10.8}{rm}A}}}}}
\put(165,275){\makebox(0,0)[lb]{\smash{{{\SetFigFont{9}{10.8}{rm}B}}}}}
\put(295,275){\makebox(0,0)[lb]{\smash{{{\SetFigFont{9}{10.8}{rm}D}}}}}
\put(320,15){\makebox(0,0)[lb]{\smash{{{\SetFigFont{9}{10.8}{rm}D}}}}}
\put(20,15){\makebox(0,0)[lb]{\smash{{{\SetFigFont{9}{10.8}{rm}A}}}}}
\put(120,15){\makebox(0,0)[lb]{\smash{{{\SetFigFont{9}{10.8}{rm}B}}}}}
\put(220,15){\makebox(0,0)[lb]{\smash{{{\SetFigFont{9}{10.8}{rm}C}}}}}
\end{picture}}
\end{center}
\caption{Linearisation of precedence ordered DAGs}
\label{fig:DAG1}
\end{figure}

\section{Linearisation of precedence ordered DAGs}
The models generated by the completeness theorem interpret (the map of)
every precedence relation $p$ as a {\em directed acyclic graph} (DAG) as
depicted in figure \ref{fig:DAG1}. However sentences in natural languages
are always totally ordered ({\em i.e.} they are strings of words). This
then raises the question:
\vspace*{-0.5ex}
\begin{itemize}
\item[] {\em Is it possible to generate linearised models}?
\end{itemize}
\vspace*{-0.5ex}
For the logic that we have described this is always possible. We only
provide a graphical argument given in figure \ref{fig:DAG1} to illustrate that
this is indeed possible.

The question that arises is then:
\vspace*{-0.5ex}
\begin{itemize}
\item[] {\em What happens when we add immediate precedence}?
\end{itemize}
\vspace*{-0.5ex}
\subsection{Problem with immediate precedence}
\begin{figure}
\begin{center}
\mbox{\setlength{\unitlength}{0.0063in}
\begin{picture}(388,415)(0,-10)
\path(25,285)(80,360)
\path(76.882,352.366)(80.000,360.000)(73.656,354.731)
\path(25,360)(80,285)
\path(73.656,290.269)(80.000,285.000)(76.882,292.634)
\path(110,180)(185,180)
\path(177.000,178.000)(185.000,180.000)(177.000,182.000)
\path(110,100)(185,100)
\path(177.000,98.000)(185.000,100.000)(177.000,102.000)
\path(110,245)(110,285)(55,285)
	(55,245)(110,245)
\path(55,245)(55,285)(0,285)
	(0,245)(55,245)
\path(55,360)(55,400)(0,400)
	(0,360)(55,360)
\path(110,360)(110,400)(55,400)
	(55,360)(110,360)
\path(55,160)(55,200)(0,200)
	(0,160)(55,160)
\path(110,160)(110,200)(55,200)
	(55,160)(110,160)
\path(110,80)(110,120)(55,120)
	(55,80)(110,80)
\path(55,80)(55,120)(0,120)
	(0,80)(55,80)
\path(240,160)(240,200)(185,200)
	(185,160)(240,160)
\path(295,160)(295,200)(240,200)
	(240,160)(295,160)
\path(295,80)(295,120)(240,120)
	(240,80)(295,80)
\path(240,80)(240,120)(185,120)
	(185,80)(240,80)
\path(78,0)(78,40)(0,40)
	(0,0)(78,0)
\path(156,0)(156,40)(78,40)
	(78,0)(156,0)
\put(75,260){\makebox(0,0)[lb]{\smash{{{\SetFigFont{9}{10.8}{rm}D}}}}}
\put(20,260){\makebox(0,0)[lb]{\smash{{{\SetFigFont{9}{10.8}{rm}C}}}}}
\put(20,375){\makebox(0,0)[lb]{\smash{{{\SetFigFont{9}{10.8}{rm}A}}}}}
\put(75,375){\makebox(0,0)[lb]{\smash{{{\SetFigFont{9}{10.8}{rm}B}}}}}
\put(20,175){\makebox(0,0)[lb]{\smash{{{\SetFigFont{9}{10.8}{rm}A}}}}}
\put(75,175){\makebox(0,0)[lb]{\smash{{{\SetFigFont{9}{10.8}{rm}B}}}}}
\put(75,95){\makebox(0,0)[lb]{\smash{{{\SetFigFont{9}{10.8}{rm}D}}}}}
\put(20,95){\makebox(0,0)[lb]{\smash{{{\SetFigFont{9}{10.8}{rm}C}}}}}
\put(205,175){\makebox(0,0)[lb]{\smash{{{\SetFigFont{9}{10.8}{rm}C}}}}}
\put(260,175){\makebox(0,0)[lb]{\smash{{{\SetFigFont{9}{10.8}{rm}D}}}}}
\put(142,315){\makebox(0,0)[lb]{\smash{{{\SetFigFont{9}{10.8}{rm}(Initial  Description)}}}}}
\put(193,15){\makebox(0,0)[lb]{\smash{{{\SetFigFont{9}{10.8}{rm}(Correct Model)}}}}}
\put(205,95){\makebox(0,0)[lb]{\smash{{{\SetFigFont{9}{10.8}{rm}A}}}}}
\put(260,95){\makebox(0,0)[lb]{\smash{{{\SetFigFont{9}{10.8}{rm}B}}}}}
\put(308,180){\makebox(0,0)[lb]{\smash{{{\SetFigFont{9}{10.8}{rm}(Incorrect }}}}}
\put(318,160){\makebox(0,0)[lb]{\smash{{{\SetFigFont{9}{10.8}{rm}Model)}}}}}
\put(308,100){\makebox(0,0)[lb]{\smash{{{\SetFigFont{9}{10.8}{rm}(Incorrect }}}}}
\put(318,80){\makebox(0,0)[lb]{\smash{{{\SetFigFont{9}{10.8}{rm}Model)}}}}}
\put(10,15){\makebox(0,0)[lb]{\smash{{{\SetFigFont{9}{10.8}{rm}A \& C}}}}}
\put(87,15){\makebox(0,0)[lb]{\smash{{{\SetFigFont{9}{10.8}{rm}B \& D}}}}}
\end{picture}}
\end{center}
\caption{Difficulty in guaranteeing linearisable models with immediate
  precedence}
\label{fig:DAG2}
\end{figure}
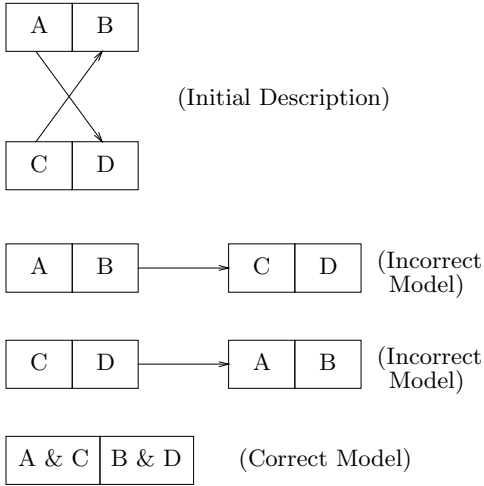

However if we add immediate precedence to our logic then it is not clear
whether we can guarantee linearisable models. This is highlighted in
figure \ref{fig:DAG2}.

As illustrated in this figure consistency checking of constraints
involving both linear precedence and immediate precedence with a
semantics that requires linearised models is  not trivial. So we do not
explore this scenario in this paper.

However, it is possible to add immediate precedence and extend the
constraint solving rules described in this paper in such a way that it is
sound and complete with respect to the current semantics described in
this paper (which does not insist on linearised models).

\section{Handling immediate precedence}
In this section, we provide additional constraint solving rules for
handling immediate precedence. The basic idea is to treat immediate
precedence as a functional relation whose inverse too is functional.

In effect what we add to our logic is both precedence as a feature and a
new constraint for representing the inverse functional precedence.

This is summarised by:
\vspace*{-1ex}
\begin{itemize}
\item {\em Represent x immediately precedes y} by :\\
  \ \hspace{4ex} $x = p: y \wedge y = p^{-1}: x$

\item Semantics:
$\begin{array}[t]{l}
  \I, \alpha \models y =  p^{-1}: x \Longleftrightarrow\\
   (p^{I})^{-1}(\alpha(y)) = \{\alpha(x)\}
\end{array}$
\end{itemize}
\vspace*{-1ex}
The additional rules given in figure below are all that
is needed to handle immediate precedence.

\begin{list}{}{\leftmargin 11ex
\labelwidth 11ex}

\item[(FeatExists)] \nfrac{$x = p: y \wedge C_{s}$}{$
             x = p: y \wedge x = \exists p: y \wedge
              C_{s}$}\\
       if $x = \exists p: y \not\in C_{s}$

\item[(ExistsTrans)] \nfrac{$x = \exists p: y \wedge C_{s}$}{$
             x = \exists p: y \wedge x = \exists p^{+}: y \wedge
              C_{s}$}\\
       if $x = \exists p^{+}: y \not\in C_{s}$

\item[(InvIntro)] \nfrac{$x = p^{-1}: y \wedge C_{s}$}{$
             y = \exists p: x \wedge x = p^{-1}: y \wedge
              C_{s}$}\\
       if $y = \exists p: x \not\in C_{s}$

\item[(InvExists)] \nfrac{$x = p^{-1}: y \wedge z = \exists p: x \wedge
C_{s}$}{$
             y = z \wedge x = p^{-1}: y \wedge y = \exists p: x \wedge
              C_{s}$}\\
       if $y \neq z$

\end{list}

\section{Conclusions}
We have shown that the logic of linear precedence can be handled
elegantly and deterministically by adding new logical primitives to
feature logic. Although, theoretically speaking, our logic comes with
some restrictions these have no practical consequences whatsoever. Our
implementation of the logic as an extension to the ProFIT typed feature
formalism shows that a reasonably efficient implementation is feasible.
Some further work is necessary to determine the computational complexity
of our constraint solving procedure. However, we believe that it is
polynomial.

The logic presented in this paper generalises the approach taken in
\cite{Reape:ThingsInOrder}. Our approach demonstrates that it is not
necessary to employ a non-deterministic operation such as {\em domain
  union} to manipulate domains. Instead precedence constraints are
directly embedded in feature logic and a deterministic constraint solving
procedure is provided. A wide range of constraints involving precedence
is provided directly in feature logic ranging from constraints expressing
{\em precedence between variables}, {\em precedence between domains} to
{\em guards on precedence constraints}.
\vspace*{-1ex}
\section{Acknowledgments}
\vspace*{-1ex}
This work was supported by The Commission of the European Communities
through the project LRE-61-061 ``Reusable Grammatical Resources'', where
the logic described in this paper has been implemented. Thanks to
Wojciech Skut for developing sample grammars to test the implementation
and for working on the interface to ProFIT. Thanks to Gregor Erbach for
demoing the extended system dubbed CL-ONE. Thanks to Herbert Ruessink and
Craig Thiersch for using and providing feedback on the implementation.
And thanks to Ralf Steinberger for providing useful comments on an
earlier draft.
\vspace*{-1ex}
{\small
\vspace*{-2ex}

}

\end{document}